\documentclass{jpsj2}

\usepackage{latexsym}
\usepackage{bm}
\usepackage{graphicx}

\def\gsize{0.55} 
\def\gs2{0.40} 

\title{Statistical Mechanics of Nonlinear 
On-line Learning for Ensemble Teachers}

\author{
\textsc{Hideto UTSUMI}$^{1}$,$^{2}$,
\textsc{Seiji MIYOSHI}$^{1}$
\thanks{E-mail address: miyoshi@kobe-kosen.ac.jp}
and 
\textsc{Masato OKADA}$^{3}$,$^{4}$
}

\inst{
$^{1}$Department of Electronic Engineering, 
Kobe City College of Technology, \\
8--3 Gakuen-higashimachi, Nishi-ku, Kobe, 651--2194 \\
$^{2}$ Department of Electrical and Electronic Engineering, 
Faculty of Engineering,
Kobe University, \\
1--1, Rokkodai, Nada-ku, Kobe 657--8501\\
$^{3}$Division of Transdisciplinary Sciences, 
Graduate School of Frontier Sciences, \\
The University of Tokyo, 
5--1--5 Kashiwanoha, Kashiwa-shi, Chiba, 277--8561 \\
$^{4}$RIKEN Brain Science Institute, 
2--1 Hirosawa, Wako-shi, Saitama, 351--0198
}

\abst{

We analyze the generalization performance of a student 
in a model composed of  nonlinear perceptrons: 
a true teacher, ensemble teachers, and the student. 
We calculate the generalization error 
of the student analytically or numerically 
using statistical mechanics
in the framework of on-line learning. 
We treat two well-known learning rules: 
Hebbian learning and perceptron learning.
As a result,
it is proven that the nonlinear model shows 
qualitatively different behaviors from the linear model. 
Moreover, it is clarified that 
Hebbian learning and perceptron learning
show qualitatively different behaviors from each other.
In Hebbian learning, we can analytically obtain the solutions.
In this case, the generalization error monotonically decreases.
The steady value of the generalization error is 
independent of the learning rate.
The larger the number of teachers is and the more variety
the ensemble teachers have, the smaller the generalization error is.
In perceptron learning, we have to numerically obtain the solutions.
In this case, the dynamical behaviors of 
the generalization error are non-monotonic.
The smaller the learning rate is, the larger the number of teachers is;
and the more variety the ensemble teachers have, 
the smaller the minimum value of the generalization error is.
}

\kword{
on-line learning, generalization error,
ensemble teachers, true teacher
}

\begin{document}
\maketitle


\section{Introduction}%
Learning is to infer the underlying rules that dominate 
data generation using observed data.
Observed data are input-output pairs from a teacher
and are called examples.
Learning can be roughly classified into batch learning and
on-line learning \cite{Saad}.
In batch learning, given examples are used more than once.
In this paradigm, a student becomes to give correct answers
after training if the student has had adequate freedom.
However, it is necessary to have a long time and 
a large memory in which to store many examples.
On the contrary, in on-line learning, examples once used are discarded.
In this case, a student cannot give correct answers 
for all examples used in training.
However, there are merits. For example,
a large memory for storing many examples isn't necessary,
and it is possible to follow a time-variant teacher. 

Recently, we \cite{Hara,PRE} 
analyzed the generalization performance
of ensemble learning
\cite{Abe,Krogh,Urbanczik}
in a framework of on-line learning
using a statistical mechanical method \cite{Saad,NishimoriE}.
Using the same method, 
we also analyzed the generalization performance
of a student supervised by a moving teacher that goes around
a true teacher\cite{JPSJ2006,JPSJ2007}.
As a result, 
it was proven that
the generalization error of a student
can be smaller than that of 
a moving teacher,
even if the student only uses examples 
from the moving teacher.
In an actual human society, a teacher
observed by a student
does not always present the correct answer.
In many cases, the teacher is learning
and continues to change.
Therefore, the analysis of such a model
is interesting for considering the 
analogies between statistical learning theories 
and an actual human society.

On the other hand, 
in most cases in an actual human society, 
a student can observe examples from two or more teachers
who differ from each other.
Therefore, we analyze the generalization performance
of such a model and discuss the 
use of imperfect teachers in this paper.
That is, we consider a true teacher and
$K$ teachers called ensemble teachers who 
exist around the true teacher.
A student uses input-output pairs from ensemble teachers
in turn or randomly.

A model in which the true teacher, the ensemble teachers and 
the student are all linear perceptrons with noise
has already been solved analytically\cite{JPSJ2006-2}．
In that case, it was proven that
when the student's learning rate satisfies $\eta <1$,
the larger the number $K$ of ensemble teachers is 
and the more variety the ensemble teachers have,
the smaller the student's generalization error is.
On the other hand, when $\eta >1$,
the properties are completely reversed.
If the variety of ensemble teachers is rich enough, 
the direction cosine between the true teacher and the 
student becomes unity in the limit of $\eta \rightarrow 0$ and
$K \rightarrow \infty$.

However, linear perceptrons are somewhat special as neural networks
or learning machines. Nonlinear perceptrons are more common
than linear ones.
Therefore, we analyze the generalization performance of a student 
in a model composed of  nonlinear perceptrons, 
a true teacher, ensemble teachers, and the student. 
We obtain order parameters and  the generalization errors
analytically or numerically 
in the framework of on-line learning using a
statistical mechanical method.
We treat two well-known learning rules: 
Hebbian learning and perceptron learning.
As a result,
it is proven that the nonlinear model shows 
qualitatively different behaviors from the linear model. 
Moreover, it is clarified that 
Hebbian learning and perceptron learning
show qualitatively different behaviors from each other.
In Hebbian learning, we can analytically obtain the solutions.
In this case, the generalization error monotonically decreases.
The steady value of the generalization error is 
independent of the learning rate $\eta$.
The larger the number $K$ of teachers is and the more variety
the ensemble teachers have, the smaller the generalization error is.
In perceptron learning, we have to numerically obtain the solutions.
In this case, the dynamical behaviors of 
the generalization error are non-monotonic.
The smaller the learning rate $\eta$ is, 
the larger the number $K$ of teachers is;
and the more variety the ensemble teachers have, 
the smaller the minimum value of the generalization error is.

\section{Model}
\begin{figure}[htbp]
\begin{center}
\includegraphics[width=\gs2\linewidth,keepaspectratio]{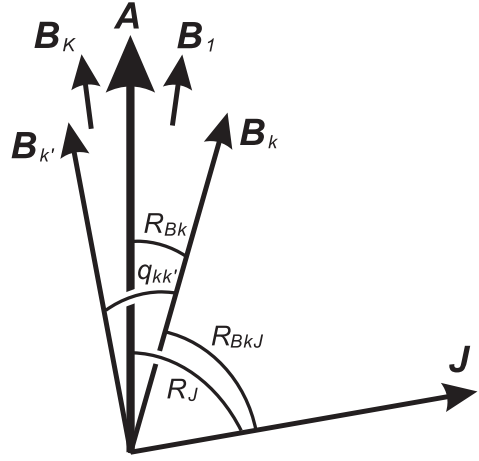}
\caption{True teacher $\mbox{\boldmath $A$}$,
ensemble teachers $\mbox{\boldmath $B$}_k$
and student $\mbox{\boldmath $J$}$.
$q_{kk'}, R_J, R_{Bk}$ and $R_{BkJ}$ are direction cosines.}
\label{fig:ABJ}
\end{center}
\end{figure}

In this paper, we consider a true teacher, $K$ ensemble teachers
and a student.
They are all nonlinear perceptrons with connection weights
$\mbox{\boldmath $A$}$, 
$\mbox{\boldmath $B$}_k$ and
$\mbox{\boldmath $J$}$, respectively.
Here, $k=1,\ldots,K$.
For simplicity, the connection weights of the true teacher,
the ensemble teachers and the student
are simply called the true teacher, the ensemble teachers and
the student, respectively.
True teacher $\mbox{\boldmath $A$}=\left(A_1,\ldots,A_N\right)$,
ensemble teachers 
$\mbox{\boldmath $B$}_k=\left(B_{k1},\ldots,B_{kN}\right)$,
student
$\mbox{\boldmath $J$}=\left(J_1,\ldots,J_N\right)$
and input 
$\mbox{\boldmath $x$}=\left(x_1,\ldots,x_N\right)$
are $N$-dimensional vectors.
Each component $A_i$ of $\mbox{\boldmath $A$}$
is drawn from ${\cal N}(0,1)$ independently and fixed,
where ${\cal N}(0,1)$ denotes Gaussian distribution with
a mean of zero and a variance of unity.
Some components $B_{ki}$
are equal to $A_i$ multiplied by --1,
and the others are equal to $A_i$.
Which component $B_{ki}$ is equal to $-A_i$
is independent of the value of $A_i$.
Hence, $B_{ki}$ also obeys ${\cal N}(0,1)$.
$B_{ki}$ is also fixed.
The direction cosine between 
$\mbox{\boldmath $B$}_k$ and
$\mbox{\boldmath $A$}$ is $R_{Bk}$
and that between
$\mbox{\boldmath $B$}_k$ and
$\mbox{\boldmath $B$}_{k'}$
is $q_{kk'}$.
Each of the components $J_i^0$ 
of the initial value $\mbox{\boldmath $J$}^0$
of $\mbox{\boldmath $J$}$
is drawn from ${\cal N}(0,1)$ independently.
The direction cosine between 
$\mbox{\boldmath $J$}$ and 
$\mbox{\boldmath $A$}$ is  $R_{J}$
and that between
$\mbox{\boldmath $J$}$ and
$\mbox{\boldmath $B$}_{k}$ is $R_{BkJ}$.
Each component $x_i$ of $\mbox{\boldmath $x$}$
is drawn from ${\cal N}(0,1/N)$ independently.
Thus,
\begin{eqnarray}
\left\langle A_i\right\rangle &=& 0, \ \ 
\left\langle \left(A_i\right)^2\right\rangle=1, \\
\left\langle B_{ki}\right\rangle &=& 0, \ \ 
\left\langle \left(B_{ki}\right)^2\right\rangle=1,\\
\left\langle J_i^0\right\rangle &=& 0, \ \ 
\left\langle \left(J_i^0\right)^2\right\rangle=1,\\
\left\langle x_i\right\rangle &=& 0, \ \ 
\left\langle \left(x_i\right)^2\right\rangle=\frac{1}{N}, \\
R_{Bk}&=&\frac{\mbox{\boldmath $A$}\cdot\mbox{\boldmath $B$}_k}{\|\mbox{\boldmath $A$}\|\|\mbox{\boldmath $B$}_k\|}, \ \ 
q_{kk'}=\frac{\mbox{\boldmath $B$}_k \cdot \mbox{\boldmath $B$}_{k'}}{\|\mbox{\boldmath $B$}_k \|\| \mbox{\boldmath $B$}_{k'}\|}, \\
R_{J}&=&\frac{\mbox{\boldmath $A$}\cdot\mbox{\boldmath $J$}}{\|\mbox{\boldmath $A$}\|\|\mbox{\boldmath $J$}\|}, \ \ 
R_{BkJ}=\frac{\mbox{\boldmath $B$}_k \cdot \mbox{\boldmath $J$}}{\|\mbox{\boldmath $B$}_k \|\| \mbox{\boldmath $J$}\|},
\end{eqnarray}
where $\langle \cdot \rangle$ denotes a mean.
Figure \ref{fig:ABJ} illustrates
the relationship among true teacher
$\mbox{\boldmath $A$}$,
ensemble teachers $\mbox{\boldmath $B$}_k$,
student $\mbox{\boldmath $J$}$
and direction cosines
$q_{kk'}, R_{Bk}, R_J$ and $R_{BkJ}$.

In this paper, the thermodynamic limit $N\rightarrow \infty$
is also treated. Therefore,
\begin{equation}
\|\mbox{\boldmath $A$}\|=\sqrt{N},\ \ 
\|\mbox{\boldmath $B$}_k\|=\sqrt{N},\ \ 
\|\mbox{\boldmath $J$}^0\|=\sqrt{N},\ \ 
\|\mbox{\boldmath $x$}\|=1.
\label{eqn:xBJ}
\end{equation}

Generally, a norm $\|\mbox{\boldmath $J$}\|$
of the student
changes as the time step proceeds.
Therefore, ratios $l^m$ of the norm to $\sqrt{N}$
are introduced and called the length of 
the student. That is,
$\|\mbox{\boldmath $J$}^m\|=l^m\sqrt{N}$,
where $m$ denotes the time step.

The internal potentials 
$y^m$ of the true teacher, 
$v_k^m$ of the ensemble teachers,
and $u^ml^m$ of the student are
\begin{eqnarray}
y^m &=&
 \mbox{\boldmath $A$}\cdot \mbox{\boldmath $x$}^m, \label{eqn:y}\\
v_k^m &=& 
 \mbox{\boldmath $B$}_k\cdot \mbox{\boldmath $x$}^m, \label{eqn:v}\\
u^m l^m &=& 
 \mbox{\boldmath $J$}^m\cdot \mbox{\boldmath $x$}^m, \label{eqn:u}
\end{eqnarray}
respectively.
Here, $y^m$, $v_k^m$ and $u^m$ 
obey the Gaussian distributions 
with means of zero and the 
covariance matrix $\mbox{\boldmath $\Sigma$}$:
\begin{eqnarray}
\mbox{\boldmath $\Sigma$}
  &=&
   \left(
   \arraycolsep=3pt
   \begin{array}{ccc}
     1       & R_{Bk}   & R_J       \\
     R_{Bk}  &   1      & R_{BkJ}    \\
     R_J     & R_{BkJ}  &    1  
   \end{array}
   \right).\label{eqn:Sigma}
\end{eqnarray}

The outputs of the true teacher, the ensemble teachers,
and the student are $\mbox{sgn}(y^m), \mbox{sgn}(v_k^m)$ 
and $\mbox{sgn}(u^m l^m)$, respectively.
Here, $\mbox{sgn}(\cdot)$ is a sign function defined as
\begin{eqnarray}
\mbox{sgn}(z)
&=&\left\{
\begin{array}{ll}
+1,            & z \geq 0 , \\
-1,            & z <    0 .
\end{array}
\right.
\label{eqn:sgn}
\end{eqnarray}

In the model treated in this paper, 
the student $\mbox{\boldmath $J$}$ is updated 
using an input $\mbox{\boldmath $x$}$ and 
the outputs of ensemble teachers $\mbox{\boldmath $B$}_k$
for the input.
That is,
\begin{eqnarray}
\mbox{\boldmath $J$}^{m+1}
&=&\mbox{\boldmath $J$}^{m}+f^{m}\mbox{\boldmath $x$}^{m},
\label{eqn:updateJ}
\end{eqnarray}
where $f^m$ denotes a function that represents the update
amount and is determined by the learning rule.
In the well-known learning rules for nonlinear perceptrons, 
Hebbian learning and  perceptron learning, 
$f^m$ are
\begin{eqnarray}
f^m &=&
 \eta \ \mbox{sgn}(v^m), \label{eqn:hebb}\\
f^m &=& 
 \eta \ \Theta (-u^m v^m) \mbox{sgn}(v^m), \label{eqn:percept}
\end{eqnarray}
respectively.
Here, $\eta$ is the learning rate of the student and is constant.
$\Theta(\cdot)$ is a step function defined as
\begin{eqnarray}
\Theta (z)
&=&\left\{
\begin{array}{ll}
+1,           & z \geq 0 , \\
0,            & z <    0 .
\end{array}
\right.\label{eqn:Theta}
\end{eqnarray}

\section{Theory}
\subsection{Generalization error}
A goal of statical learning theory is to theoretically obtain
generalization errors.
We use
\begin{equation}
\epsilon^m=\Theta(-y^m u^m)
\end{equation}
as the error of the student. 
The superscripts $m$, which represent the time step, are
omitted for simplicity unless stated otherwise.
Since the generalization error is the mean of errors for the true
teacher over the distribution of new input, generalization
error $\epsilon_{g}$ of student $\mbox{\boldmath $J$}$ is
calculated as follows:
\begin{eqnarray}
\epsilon_{g} 
&=& \int d\mbox{\boldmath $x$}P(\mbox{\boldmath $x$})\epsilon \\
&=& \int dyduP(y,u)\epsilon(y,u) \\
&=&  \frac{1}{\pi}\tan^{-1}{\frac{\sqrt{1-{R_J}^2}}{R_J}}.
\label{eqn:eg}
\end{eqnarray}
Here, integration has been executed using the following: 
$y$ and $u$ obey ${\cal N}(0,1)$. 
The covariance between $y$ and $u$ is $R_J$.

\subsection{Differential equations for order parameters}
To simplify the analysis, the following auxiliary order parameters
are introduced:
\begin{eqnarray}
r_J      &\equiv& R_J     l, \label{eqn:defrJ} \\
r_{BkJ}  &\equiv& R_{BkJ} l. \label{eqn:defrBkJ}
\end{eqnarray}

Simultaneous differential equations
in deterministic forms \cite{NishimoriE}, 
which describe the dynamical behaviors of order parameters,
have been obtained based on self-averaging
in the thermodynamic limits as follows:
\begin{align}
\frac{dr_{BkJ}}{dt}&= 
\frac{1}{K}\sum_{k'=1}^K \langle f_{k'} v_k\rangle, \label{eqn:drBkJdt}\\
\frac{dr_J}{dt}&=
\frac{1}{K}\sum_{k=1}^K \langle f_k y\rangle, \label{eqn:drJdt}\\
\frac{dl}{dt}&=
\frac{1}{K}\sum_{k=1}^K
\left(\langle f_k u\rangle+\frac{1}{2l}\langle f_k^2 \rangle \right).
\label{eqn:dldt}
\end{align}
Here, dimension $N$ has been treated 
to be sufficiently greater than
the number $K$ of ensemble teachers.
Time is defined by $t=m/N$, that is, 
time step $m$ normalized by dimension $N$.
Note that the above differential equations
are identical whether 
the $K$ ensemble teachers
are used in turn or randomly.

\subsection{Hebbian learning}
Since $y$, $v$ and $u$
obey the triple Gaussian distribution
with means of zero and the covariance matrix
of eq. (\ref{eqn:Sigma}),
the four sample averages that appear in
eqs. (\ref{eqn:drBkJdt})--(\ref{eqn:dldt}) 
in Hebbian learning
can be calculated
using eq.(\ref{eqn:hebb}) as follows:
\begin{align}
\langle f_{k'} v_k \rangle
&= \eta \frac{2q_{kk'}}{\sqrt{2\pi}}, \label{eqn:hfk'vk} \\
\langle f_k y \rangle 
&= \eta \frac{2R_{Bk}}{\sqrt{2\pi}}, \label{eqn:hfky} \\
\langle f_k u \rangle 
&= \eta \frac{2R_{BkJ}}{\sqrt{2\pi}}, \label{eqn:hfku} \\
\langle f_k^2 \rangle
&= \eta^2. \label{eqn:hfk2} 
\end{align}

Since all components $A_i$, $J_i^0$ 
of true teacher $\mbox{\boldmath $A$}$,
and the initial student $\mbox{\boldmath $J$}^0$
are drawn from ${\cal N}(0,1)$ independently
and because the thermodynamic limit $N\rightarrow \infty$
is also treated,
they are orthogonal to each other in the initial state.
That is,
\begin{equation}
R_J=0\ \mbox{when}\ t=0.
\label{eqn:Rinit}
\end{equation}

In addition,
\begin{equation}
l=1\ \mbox{when}\ t=0.
\label{eqn:linit}
\end{equation}

Using eqs. (\ref{eqn:hfk'vk})--(\ref{eqn:linit}), the simultaneous 
differential equations 
(\ref{eqn:drBkJdt})--(\ref{eqn:dldt}) can be solved 
analytically as follows: 
\begin{eqnarray}
r_{BkJ} &=& 
\frac{\eta}{K}\sum_{k'=1}^K \frac{2q_{kk'}}{\sqrt{2\pi}}t, 
\label{eqn:rBkJ} \\
r_J &=& \frac{\eta}{K}\sum_{k=1}^K \frac{2R_{Bk}}{\sqrt{2\pi}}t, 
\label{eqn:rJ}\\
l^2 &=&
\frac{\eta^2}{K}\sum_{k=1}^K
\left(\frac{2}{K\pi}\sum_{k'=1}^K q_{kk'}t^2+t\right)+1. 
\label{eqn:l2}
\end{eqnarray}

\subsection{Perceptron learning}
Since $y$, $v$ and $u$
obey the triple Gaussian distribution
with means of zero and the covariance matrix
of eq. (\ref{eqn:Sigma}),
the four sample averages that appear in
eqs. (\ref{eqn:drBkJdt})--(\ref{eqn:dldt}) 
in perceptron learning
can be calculated
using eq. (\ref{eqn:percept}) as follows:
\begin{align}
\langle f_{k'}v_k \rangle
&= \eta \frac{q_{kk'}-R_{BkJ}}{\sqrt{2\pi}}, \label{eqn:pfk'vk} \\
\langle f_k y \rangle 
&= \eta \frac{R_{Bk}-R_J}{\sqrt{2\pi}}, \label{eqn:pfky}\\  
\langle f_k u \rangle
&= \eta \frac{R_{BkJ}-1}{\sqrt{2\pi}}, \label{eqn:pfku}\\
\langle f_k^2 \rangle
&= \frac{\eta^2}{\pi}\tan^{-1}\frac{\sqrt{1-R_{BkJ}^2}}{R_{BkJ}}. 
\label{eqn:pfk2}
\end{align}

Since the simultaneous differential equations
cannot be solved analytically in this case, 
we solve these equations numerically. 

\section{Results and Discussion}
In this section, 
we treat the case where the direction cosines $R_{Bk}$ between 
the ensemble teachers and the true teacher,
and the direction cosines $q_{kk'}$ among the ensemble teachers
are uniform.
That is,
\begin{eqnarray}
R_{Bk}         &=& R_B,\ \  k=1,\ldots,K, \label{eqn:RBk}\\
q_{kk'}
&=&\left\{
\begin{array}{ll}
q,           & k\neq k' , \\
1,           & k=k'.
\end{array}
\right.\label{eqn:qkk} 
\end{eqnarray}

In Hebbian learning, since order parameters are analytically
obtained,
we can understand the dynamical behaviors clearly and deeply.
Considering eqs. (\ref{eqn:defrJ}), (\ref{eqn:rJ}), 
(\ref{eqn:l2}), (\ref{eqn:RBk}) and (\ref{eqn:qkk})，
$R_J$ is obtained as follows:
\begin{equation}
R_J = 
\frac{R_B}{\sqrt{\frac{(K-1)q+1}{K}+
\frac{\pi}{2}\left(\frac{1}{\eta^2 t^2}+\frac{1}{t}\right)}}.
\label{eqn:RJ-Miyo}
\end{equation}

Equation (\ref{eqn:RJ-Miyo}) shows the following: 
the dynamical behaviors of $R_J$ are monotonically increasing.
The larger the learning rate $\eta$ is, the larger 
the direction cosine $R_J$ is.
$R_J$ in the limit of $t \rightarrow \infty$ is 
obtained as follows:
\begin{eqnarray}
R_J &\rightarrow& \frac{R_B}{\sqrt{\frac{1}{K}+(1-\frac{1}{K})q}}
= \frac{R_B}{\sqrt{q+\frac{1-q}{K}}}. \label{eqn:stlRJ2}
\end{eqnarray}

This equation shows that 
the steady state value of $R_J$ is
independent of the learning rate $\eta$.
The larger the number $K$ of ensemble teachers is 
and the smaller the direction cosine $q$ among ensemble teachers is,
the larger the steady state value of $R_J$ is.

Considering that 
the generalization error $\epsilon_g$ calculated
by eq.(\ref{eqn:eg})
monotonically decreases as $R_J$ increases,
$\epsilon_g$ in the case of Hebbian learning
monotonically decreases.
The larger $\eta$ is, the smaller $\epsilon_g$ is
in the transient phase.
The steady state value of $\epsilon_g$ is independent of $\eta$.
However, 
the larger the number $K$ is and the smaller $q$ is,
the smaller the steady state value of $\epsilon_g$ is.
Therefore, the larger the number of teachers is
and the more variety the ensemble teachers have, 
the more clever the student can become.

\begin{figure}[htbp]
\begin{center}
 \includegraphics[width=\gsize\linewidth,keepaspectratio]{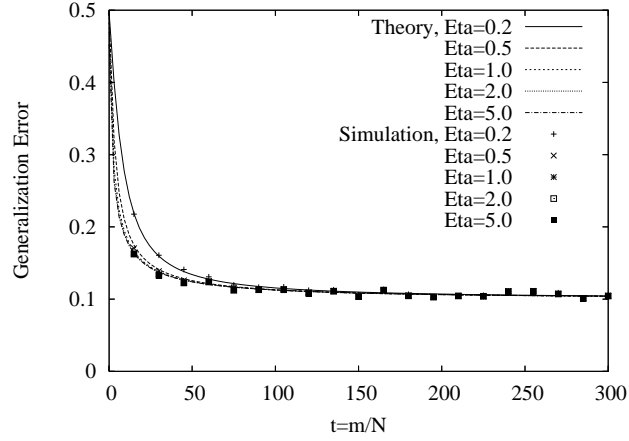}
\caption{Dynamical behaviors of generalization error $\epsilon_g$.
Hebbian learning.
Theory and computer simulations.
Conditions other than $\eta$ are
$K=10, q=0.49$ and $R_B=0.7$.}
\label{fig:H-t-eta}
\end{center}
\end{figure}

\begin{figure}[htbp]
\begin{center}
\includegraphics[width=\gsize\linewidth,keepaspectratio]{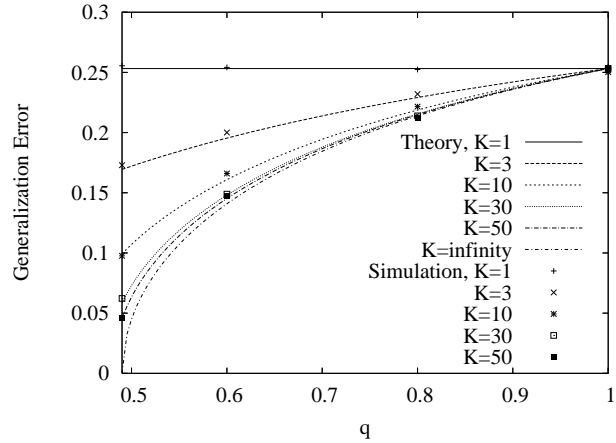}
\caption{Steady state value of  generalization error
$\epsilon_{g}$.
Hebbian learning.
Theory and computer simulations. $R_B=0.7$.
When $q=R_B{}^2$ and $K=\infty$, the steady state value of 
$\epsilon_{g}$ is zero.}
\label{fig:H-q-K}
\end{center}
\end{figure}

Equation (\ref{eqn:stlRJ2}) shows 
$R_J \rightarrow R_B/\sqrt{q}$ in the limit of 
$K\rightarrow \infty$.
On the other hand,
when $\mbox{\boldmath $S$}$ and 
$\mbox{\boldmath $T$}$ are generated independently
under conditions where the direction cosine 
between $\mbox{\boldmath $S$}$ and $\mbox{\boldmath $P$}$
and 
between $\mbox{\boldmath $T$}$ and $\mbox{\boldmath $P$}$
are both $R_0$, where $\mbox{\boldmath $S$}$, 
$\mbox{\boldmath $T$}$ and $\mbox{\boldmath $P$}$
are high dimensional vectors,
the direction cosine between 
$\mbox{\boldmath $S$}$ and $\mbox{\boldmath $T$}$
is $q_0=R_0^2$, as shown in the appendix.
Therefore, 
if ensemble teachers have enough variety
that they have been generated independently
under the condition that all direction cosines
between ensemble teachers and the true teacher are
$R_B$, $R_B/\sqrt{q}=1$,
then the direction cosine $R_J$ between
the student and the true teacher approaches unity
in the limit of 
$K \rightarrow \infty$.
Then, the generalization error approaches zero.

The dynamical behaviors of generalization error
$\epsilon_g$
have been analytically obtained by 
eqs.(\ref{eqn:eg}) and (\ref{eqn:RJ-Miyo})
in Hebbian learning.
Figures \ref{fig:H-t-eta} and \ref{fig:H-q-K} show
the analytical results of $\epsilon_{g}$ and 
the steady state value of $\epsilon_{g}$
with corresponding simulation results.
In computer simulations, 
the dimension $N=2000$ and 
$K$ ensemble teachers are used in turn.
The generalization error $\epsilon_{g}$
was obtained by
test for $10^4$ random inputs at each time step.
In these figures, 
the curves represent theoretical results. 
The symbols 
represent simulation results.
In Fig. \ref{fig:H-t-eta}, conditions other than $\eta$
are common:
$K=10, q=0.49$ and $R_B=0.7$.
In Fig. \ref{fig:H-q-K}, only $R_B$ is common: $R_B =0.7$.
The former discussions are confirmed in these figures.

\begin{figure}[htbp]
\begin{center}
\includegraphics[width=\gsize\linewidth,keepaspectratio]{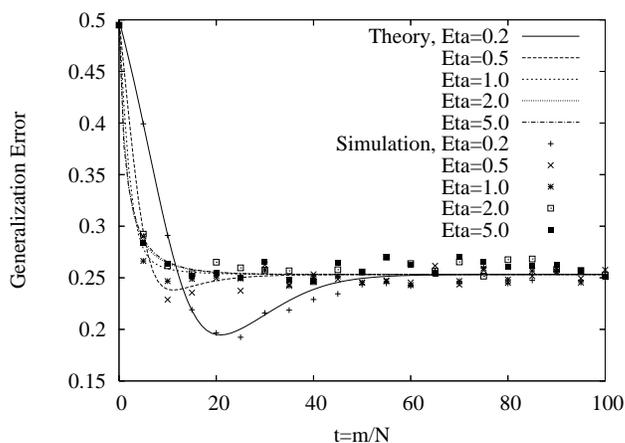}
\caption{Dynamical behaviors of generalization error
$\epsilon_g$.
Perceptron learning.
Theory and computer simulations.
Conditions other than $\eta$ are
$K=10, q=0.49, R_B=0.7$.}
\label{fig:P-t-eta}
\end{center}
\end{figure}

\begin{figure}[htbp]
\begin{center}
\includegraphics[width=\gsize\linewidth,keepaspectratio]{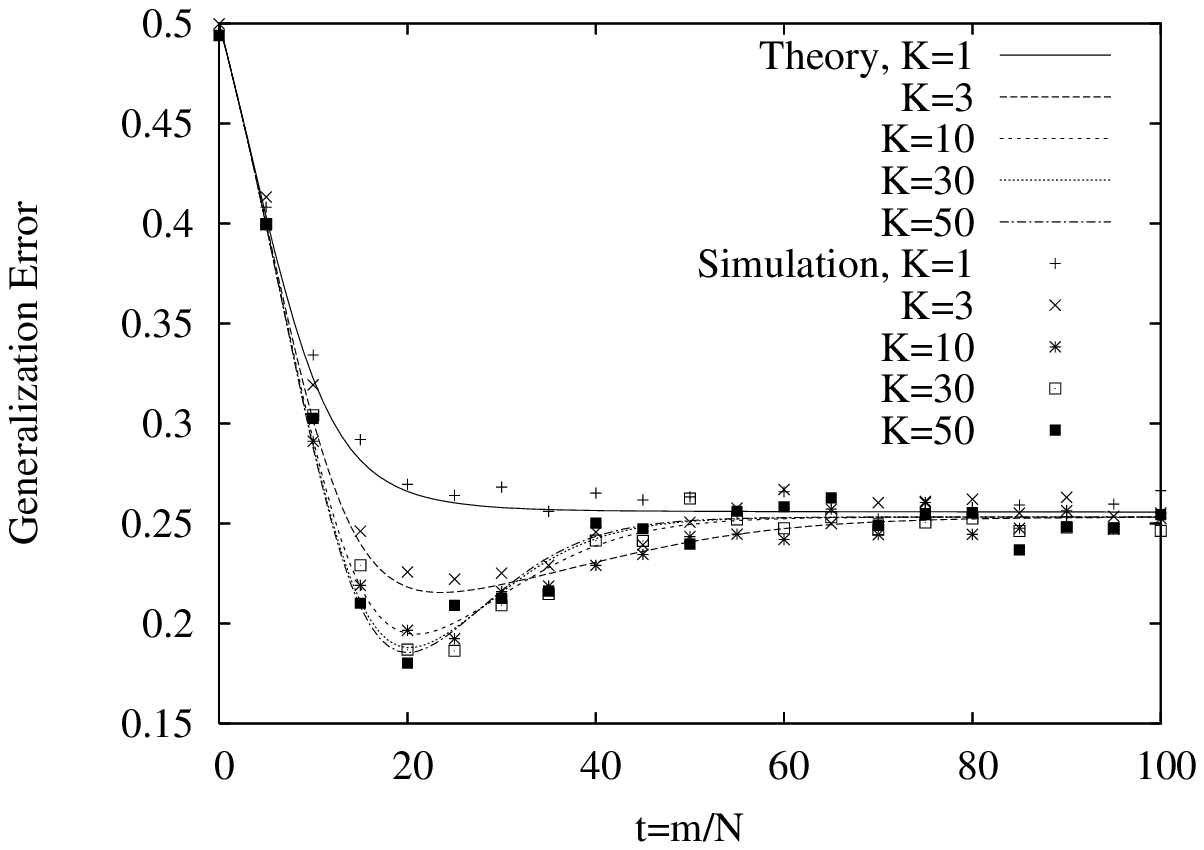}
\caption{Dynamical behaviors of generalization error 
$\epsilon_g$.
Perceptron learning.
Theory and computer simulations.
Conditions other than $K$ are
$\eta=0.2, q=0.49$ and $R_B=0.7$.}
\label{fig:P-t-K}
\end{center}
\end{figure}

\begin{figure}[htbp]
\begin{center}
\includegraphics[width=\gsize\linewidth,keepaspectratio]{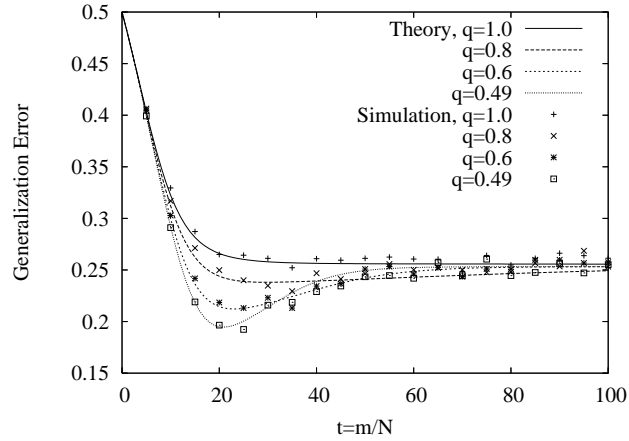}
\caption{Dynamical behaviors of generalization error 
$\epsilon_g$.
Perceptron learning.
Theory and computer simulations.
Conditions other than $q$ are
$K=10, \eta=0.2$ and $R_B=0.7$.}
\label{fig:P-t-q}
\end{center}
\end{figure}

On the other hand, in perceptron learning,
we cannot solve 
eqs.(\ref{eqn:drBkJdt})--(\ref{eqn:dldt})
analytically.
Therefore, we obtain the solutions numerically.
The dynamical behaviors of 
generalization errors $\epsilon_g$ 
are shown in Figs. \ref{fig:P-t-eta}--\ref{fig:P-t-q}.

In Fig.\ref{fig:P-t-eta}, conditions other than $\eta$ are
$K=10, q=0.49$ and $R_B=0.7$.
In Fig.\ref{fig:P-t-K}, conditions other than $K$ are
$\eta=0.2, q=0.49$ and $R_B=0.7$.
In Fig.\ref{fig:P-t-q}, conditions other than $q$ are
$K=10, \eta=0.2$ and $R_B=0.7$.
Figure \ref{fig:P-t-eta} shows that 
the dynamical behaviors of $\epsilon_g$ have
non-monotonic properties when the learning rate $\eta$ is
relatively small.
However, Figs.\ref{fig:P-t-K} and \ref{fig:P-t-q}
show that the steady state value of the generalization error
is independent of $K$ and $q$.
These are remarkable differences from 
the properties of Hebbian learning.

\begin{figure}[htbp]
\begin{center}
\includegraphics[width=\gsize\linewidth,keepaspectratio]{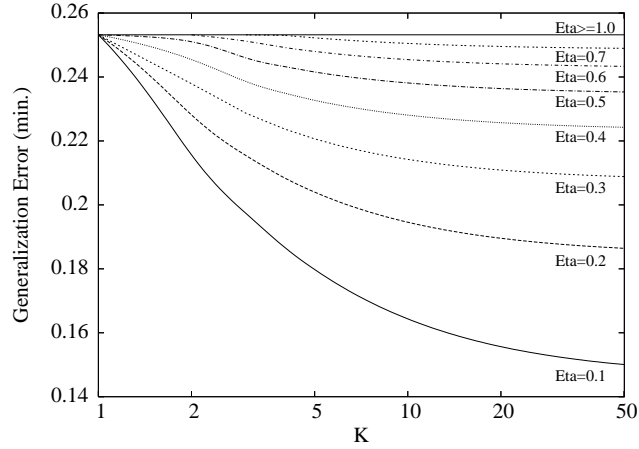}
\caption{Relationship between $K$ and 
minimum values $\epsilon_{g}(\mbox{min})$ of generalization error.
Perceptron learning.
Theory.
$q=0.49, R_B=0.7$.}
\label{fig:egmin-K-eta}
\end{center}
\end{figure}

\begin{figure}[htbp]
\begin{center}
\includegraphics[width=\gsize\linewidth,keepaspectratio]{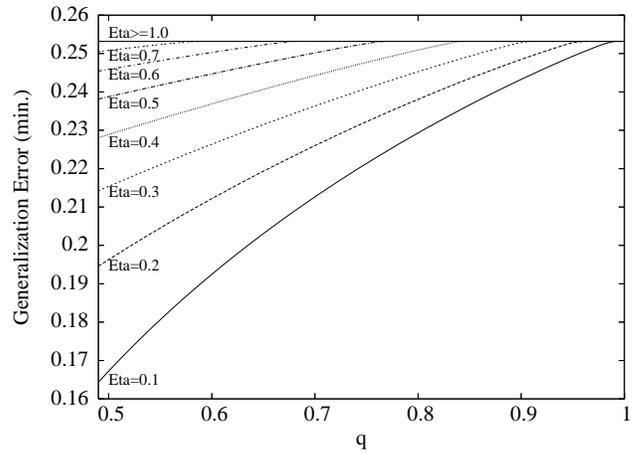}
\caption{Relationship between $q$ and 
minimum values $\epsilon_{g}(\mbox{min})$ of generalization error.
Perceptron learning.
Theory. $K=10, R_B=0.7$.}
\label{fig:egmin-q-eta}
\end{center}
\end{figure}

When the learning rate $\eta$ is relatively small, 
the minimum value $\epsilon_{g}(\mbox{min})$ of 
the generalization error
exists and the smaller $\eta$ is, 
the smaller $\epsilon_{g}(\mbox{min})$ is. 
The relationships between $K$ and $\epsilon_{g}(\mbox{min})$,
and $q$ and $\epsilon_{g}(\mbox{min})$ are shown in
Figs. \ref{fig:egmin-K-eta} and \ref{fig:egmin-q-eta},
respectively.
In Fig.\ref{fig:egmin-K-eta}, conditions other than $\eta$ are
$q=0.49$ and $R_B=0.7$.
In Fig.\ref{fig:egmin-q-eta}, conditions other than $\eta$ are
$K=10$ and $R_B=0.7$.
These figures show that
the larger the number $K$ is
and the smaller the direction cosine $q$ is,
the smaller the minimum value of generalization errors is.
In other words, the larger the number of teachers is
and the more variety the ensemble teachers have, 
the more clever the student can become.

In the case of the linear model\cite{JPSJ2006-2},
the properties were able to be summarized as follows:
The smaller $\eta$ is, the smaller the steady state value 
of $\epsilon_g$ is.
When the learning rate satisfies $\eta<1$, 
the larger $K$ is and the smaller $q$ is,
the smaller the steady state value of $\epsilon_e$ is. 
On the contrary, when $\eta>1$, 
the properties are completely reversed\cite{JPSJ2006-2}.
Comparing the linear model and the nonlinear model
treated in this paper,
there are qualitatively different properties.

\section{Conclusion}
We have analyzed the generalization performance of a student 
in a model composed of  nonlinear perceptrons: 
a true teacher, ensemble teachers, and the student. 
We have calculated the generalization error 
of the student analytically or numerically 
using statistical mechanics
in the framework of on-line learning. 
We have treated two well-known learning rules: 
Hebbian learning and perceptron learning.
As a result,
it has been proven that the nonlinear model shows 
qualitatively different behaviors from the linear model. 
Moreover, it has been clarified that 
Hebbian learning and perceptron learning
show qualitatively different behaviors from each other.
In Hebbian learning, we have analytically obtained the solutions.
In this case, the generalization error monotonically decreases.
The steady value of the generalization error is 
independent of the learning rate.
The larger the number of teachers is and the more variety
the ensemble teachers have, the smaller the generalization error is.
In perceptron learning, we have obtained the solutions numerically.
In this case, the dynamical behaviors of 
the generalization error are non-monotonic.
The smaller the learning rate is, the larger the number of teachers is,
and the more variety the ensemble teachers have, 
the smaller the minimum value of the generalization error is.

\section*{Acknowledgments}
This research was partially supported by the Ministry of Education, 
Culture, Sports, Science, and Technology of Japan, 
with Grants-in-Aid for Scientific Research
15500151, 16500093, 18020007, 18079003 and 18500183.

\appendix
\section{Direction cosine $q$ among ensemble teachers}
\begin{figure}[htbp]
\vspace{3mm}
\begin{center}
\includegraphics[width=\gs2\linewidth,keepaspectratio]{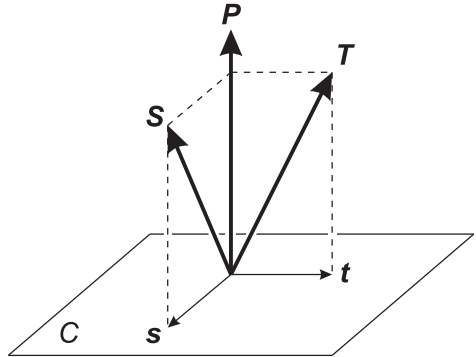}
\caption{Direction cosine among ensemble teachers.}
\label{fig:PST}
\end{center}
\end{figure}

Let us consider the case where
$\mbox{\boldmath $S$}$ and 
$\mbox{\boldmath $T$}$ are generated independently,
satisfying the condition that direction cosines 
between $\mbox{\boldmath $S$}$ and $\mbox{\boldmath $P$}$
and 
between $\mbox{\boldmath $T$}$ and $\mbox{\boldmath $P$}$
are both $R_0$,
as shown in Fig. \ref{fig:PST}, 
where $\mbox{\boldmath $S$}$, 
$\mbox{\boldmath $T$}$ and $\mbox{\boldmath $P$}$
are $N$ dimensional vectors.
In this figure, 
the inner product of 
$\mbox{\boldmath $s$}$ and
$\mbox{\boldmath $t$}$ is
\begin{eqnarray}
\mbox{\boldmath $s$}\cdot\mbox{\boldmath $t$}
&=& 
\left(\mbox{\boldmath $S$}-
R_0
\frac{\|\mbox{\boldmath $S$}\|}{\|\mbox{\boldmath $P$}\|}
\mbox{\boldmath $P$}\right)
\cdot
\left(\mbox{\boldmath $T$}-
R_0
\frac{\|\mbox{\boldmath $T$}\|}{\|\mbox{\boldmath $P$}\|}
\mbox{\boldmath $P$}\right)\\
&=&
\|\mbox{\boldmath $S$}\|\|\mbox{\boldmath $T$}\|
\left(q_0-R_0^2\right),
\end{eqnarray}
where $\mbox{\boldmath $s$}$ and $\mbox{\boldmath $t$}$ 
are
projections from 
$\mbox{\boldmath $S$}$ to
the orthogonal complement $C$ of $\mbox{\boldmath $X$}$
and 
from 
$\mbox{\boldmath $T$}$ to $C$, respectively.
$q_0$ denotes the direction cosine between
$\mbox{\boldmath $S$}$ and $\mbox{\boldmath $T$}$.

Incidentally, 
if dimension $N$ is large and 
$\mbox{\boldmath $S$}$ and $\mbox{\boldmath $T$}$
have been generated independently,
$\mbox{\boldmath $s$}$ and $\mbox{\boldmath $t$}$
should be orthogonal to each other.
Therefore, $q_0=R_0^2$.

\end{document}